# Design and Control of a Novel Variable Stiffness Series Elastic Actuator

Emre Sariyildiz, *Senior Member, IEEE*, Rahim Mutlu, *Member IEEE,* Jon Roberts, Chin-Hsing Kuo, Barkan Ugurlu, *Member IEEE*

*Abstract*— This paper expounds the design and control of a new Variable Stiffness Series Elastic Actuator (VSSEA). It is established by employing a modular mechanical design approach that allows us to effectively optimise the stiffness modulation characteristics and power density of the actuator. The proposed VSSEA possesses the following features: i) no limitation in the work-range of output link, ii) a wide range of stiffness modulation (~20Nm/rad to ~1KNm/rad), iii) low-energy-cost stiffness modulation at equilibrium and non-equilibrium positions, iv) compact design and high torque density (~36Nm/kg), and v) high-speed stiffness modulation (~3000Nm/rad/s). Such features can help boost the safety and performance of many advanced robotic systems, e.g., a cobot that physically interacts with unstructured environments and an exoskeleton that provides physical assistance to human users. These features can also enable us to utilise variable stiffness property to attain various regulation and trajectory tracking control tasks only by employing conventional controllers, eliminating the need for synthesising complex motion control systems in compliant actuation. To this end, it is experimentally demonstrated that the proposed VSSEA is capable of precisely tracking desired position and force control references through the use of conventional Proportional-Integral-Derivative (PID) controllers.

## I. INTRODUCTION

To BOOST safety in physical-robot environment interaction, compliant actuators have been widely adopted by many different advanced robotic systems such as humanoids, cobots, quadrupeds, and exoskeletons [1–5]. A compliant actuation system could be developed by simply integrating an elastic element into the design of an actuator [5]. For example, Series Elastic Actuators (SEAs), one of the most popular compliant actuation systems in robotics, are developed by placing a spring between a conventional rigid actuator and link [6–8]. In addition to improving safety, the spring between the rigid actuator and link can provide several benefits such as low-cost and high-fidelity force control, mechanical energy storage, lower reflected inertia, higher tolerance to impact loads, and increased output power [8].

Despite the aforementioned benefits, the elastic elements integrated to compliant actuators introduce certain challenges and fundamental limitations in motion control [9]. For example, it is a well-known fact that the position control problem of a compliant actuator is more complicated than that of a conventional rigid actuator [10–14]. To suppress the vibrations and disturbances of a compliant actuator's link, researchers generally need to employ advanced motion controllers [14]. Another example is that although using a softer elastic element improves safety and transparency, the natural frequency of the compliant actuator decreases. This not only excites the vibrations at link side but also lowers the bandwidth of the actuator, thus limiting achievable position and force control performance in motion control applications [13, 14]. It is therefore essential to properly choose the stiffness of the elastic element of a compliant actuator based on the target control task [15]. However, this is mostly impractical for compliant actuators with fixed elastic elements, which often leads to compromise between safety and performance [10, 16]. A simple yet efficient solution for this fundamental problem could be achieved by integrating a compliant mechanical component with variable stiffness property to actuators in series or parallel [17, 18].

Adaptable compliance mechanisms that can alter the stiffness of actuators mechanically have been developed to meet the different compliance requirements of motion control tasks such as soft actuation in human-robot interaction and stiff actuation in trajectory tracking. Although no standard terminology exists, such actuators are generally called Variable Stiffness Actuators (VSAs) in the literature. A comprehensive survey on the design of VSAs can be found in [18, 19]. Among them, antagonistic actuation is one of the most well-known and widely used stiffness modulation methods in VSAs. Inspired by mammalian anatomy, this actuation method has been studied since early 1980s [20], and different antagonistic actuators have been developed and used in various robotic applications, such as legged locomotion and upper-limb rehabilitation, since late 1990s [21, 22]. The simplest antagonistic actuator can be designed by using an agonistic-antagonistic setup which connects two motors to a link via two nonlinear springs, similar to biceps and triceps in the human arm [19, 23]. While the equilibrium position can be adjusted by rotating two motors in the same direction, counter-rotation of motors alters the

Manuscript received…..... (Corresponding author: Emre Sariyildiz).

E. Sariyildiz, J. Roberts and C.-H. Kuo are with the School of Mechanical, Materials, Mechatronic and Biomedical Engineering, University of Wollongong, Wollongong, NSW, 2522, Australia. (e-mails: emre@uow.edu.au, robertsj@uow.edu.au, chkuo@uow.edu.au).

R. Mutlu is with the Faculty of Engineering and Information Sciences, University of Wollongong in Dubai, Dubai, United Arab Emirates and the Intelligent Robotics & Autonomous Systems Co (iR@SC), NSW, 2529, Australia. (e-mail: ramutlu@irasc.com.au).

B. Ugurlu is with the Department of Mechanical Engineering, Ozyegin University, Istanbul, 34794, Turkey. (e-mail: barkan.ugurlu@ozyegin.edu.tr).

stiffness of the actuator by changing the spring preload. Despite its simplicity, this bioinspired actuation method has several drawbacks in practice, e.g., i) the control problems of stiffness modulation and equilibrium position are coupled, ii) the output torque of the actuator is limited by the maximum torque of each motor, iii) the potential energy capacity of nonlinear springs cannot be used entirely, and iv) the energy consumption is high because preloading nonlinear springs for stiffness modulation requires constant power drain, even when the actuator does not perform net mechanical work at equilibrium positions [24, 25].

Numerous different antagonistic and non-antagonistic VSAs have been proposed to tackle the drawbacks of the agonistic-antagonistic actuation setup in the last two decades [18, 19]. For example, while the output torque of antagonistic actuators is increased using bi-directional configuration in [26], a partially decoupled motion control problem is obtained using a quasi-antagonistic actuation mechanism in [27]. However, the aforementioned problems could not be entirely addressed using antagonistic actuation systems [18, 19]. This has motivated many researchers to build non-antagonistic VSAs. The control problems of the equilibrium position and compliance of the actuator's output link are decoupled using a new mechanical design approach in Maccepa [28]. The stiffness of the actuator is modulated by controlling the tension of a linear spring. A similar stiffness modulation approach is employed in DLR-VSJ, and a light-weight and compact VSA is designed by changing only the cam disk of the joint in [29]. Nevertheless, similar to antagonistic actuators, high-energy-cost stiffness modulation remains a challenging problem in these non-antagonistic VSA design approaches. Since the stiffness of the Maccepa and DLR-VSJ is modulated by pretensioning springs, they require constant power drain at equilibrium positions, thus leading to high energy consumption [28 – 30]. The energy-cost of stiffness modulation has been improved using different VSA design approaches in the last decade. The stiffness of the output link is modulated by changing the positions of the springs and pivot points on a lever arm in AWAS [31, 32]. Low energy cost stiffness modulation (e.g., theoretically zero power drain at equilibrium positions) could be achieved using this VSA design approach. The stiffness range, however, is limited by the size of the actuator [31, 33]. In vsaUT, the stiffness of the actuator is modulated by controlling the effective length of a lever arm [34, 35]. This stiffness modulation approach allows us to attain not only zero power drain at equilibrium positions but also infinite-range stiffness modulation with compact actuators. However, the power drain by the motor dedicated to stiffness modulation becomes unbounded as the stiffness of the actuator approaches infinity [33, 34]. Variable length leaf spring mechanisms have also been employed to develop energy efficient VSAs [30, 36 – 39]. Compared to the other antagonistic and non-antagonistic VSAs, recent studies show that variable length leaf spring mechanisms can provide several benefits in practice, e.g., zero power drain at equilibrium positions, fast and infinite-range stiffness modulation, and bounded power drain for all stiffness ranges at equilibrium and non-equilibrium positions [30, 33]. These features could be very useful in biomedical engineering applications as shown in [30, 38]. Nevertheless, when it comes to building a compact VSA that can be integrated to different robotic systems such as cobots, the variable length leaf spring mechanisms may involve several drawbacks such as low torque/power density and work-space limitations [30, 36 – 39].

It is noted that a non-antagonistic VSA can be simply built by employing a discrete stiffness modulation method where multiple springs could be integrated to actuators in series or parallel [40 – 42]. The main drawback of this actuation method is the limited stiffness range which depends on the number of springs employed in the actuator design. Moreover, the discrete stiffness modulation method leads to several challenges in controller analysis and synthesis such as the stability problem of switching systems [40, 42]. Therefore, continuous stiffness modulation methods are mainly considered in this paper.

The existing VSAs have their own merits and demerits. While they are highly functional in their own domain of use, they may however fall-short in complying with all the desirable technical specifications of an ideal compliant actuator for practical applications: i) a compact and simple mechanical design that allows to easily reconfigure a VSA for different applications, ii) no limitation in motion range, iii) a wide range of stiffness modulation, iv) rapid stiffness change, and v) energy efficient actuation [30]. In general, researchers manage a trade-off to target only few of the aforementioned qualities, leading to different compromises such as high energy consumption or limited motion control performance in robotic applications. Hence, despite many recent advances, more effort should be put into the development of VSAs [18, 19 and 30]. This is summarised using the examples of existing VSAs in the literature in Table I.

To this end, this paper proposes a new VSSEA which consists of three main components: i) a rigid actuator that independently controls the equilibrium position, ii) a novel Variable Stiffness Actuation Mechanism (VSAM), and iii) a direct drive motor that independently adjusts the stiffness. The proposed VSSEA

TABLE I: Actuators, Deflection Range, Motion Range, Stiffness Range, Energy Cost of Fixed Stiffness at Equilibrium, Energy Cost of Infinite-Range Stiffness Modulation, Torque Density and Stiffness Modulation Speed.

| Actuators | Deflection Range [degree] | Motion Range [degree] | Stiffness range [Nm/rad] | Energy cost of fixed stiffness at equilibrium | Energy cost of infinite-range stiffness modulation | Torque density [Nm/kg] | Stiffness modulation speed [Nm/rad/s] |
|---|---|---|---|---|---|---|---|
| VSSEA | ±25 | no limitation | 20 – 1000 [b] | ZEC | BEC | 35.72 | 3000 |
| Maccepa | ±60 | ±90 [c] | 5 – 110 | NZEC | NA | 20.83 | 40 |
| DLR-VSJ | ±15 | ±180 | 50 – 820 | NZEC | NA | 22.27 | 2350 |
| AWAS | ±12 | ±120 | 30 – 1500 | ZEC | NA | 31.1 | 420 |
| AWAS-II | ±18 | ±150 | 10 – 10000 | ZEC | NA | 9.76 | 4000 |
| vsaUT-II | ±45 | ±180 [a] | 0.5N – 100 | ZEC | NBEC | 8.7 | 1000 |
| VSA-I [36] | ±12 | NA | 250 – 3000 [b] | ZEC | BEC | 6.1 | NA |
| VSA-II [30] | ±75 | ±75 | 10 – 8000 [b] | ZEC | BEC | Passive (3kg) | 10000 |

Table I is prepared using experimental results given in the papers unless data is provided in the references. [a] Although the motion range of vsaUT is limited (e.g., vsaUT-II's motion range is ±180°), mvsaUT can perform continuous motions without any motion range limitations [46]. [b] Infinite-range stiffness modulation can be achieved using leaf spring based VSAs. [c] [28] states that the motion range of Maccepa can be increased. ZEC: Zero Energy Consumption; NZEC: Non-Zero Energy Consumption; BEC: Bounded Energy Consumption when performing infinite-range stiffness modulation; NBEC: Non-Bounded Energy Consumption when performing infinite-range stiffness modulation; and NA: Not Applicable because the actuators cannot provide infinite range stiffness modulation.

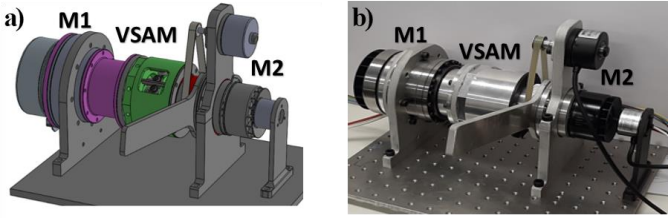

a) CAD model. b) First prototype of the VSSEA.
Figure 1: CAD model and first prototype of the novel VSSEA. M1: Motor 1, M2: Motor 2, and VSAM: Variable Stiffness Actuation Mechanism.

is simply developed by integrating the VSAM to a rigid actuator. This modular design approach allows us to easily optimise the stiffness modulation characteristics and output power/torque of the VSSEA for different robotic applications. The stiffness of the actuator is modulated by changing the effective length of a group of leaf springs of the VSAM through a direct drive motor. This stiffness modulation technique provides several benefits: i) stiffness modulation over a large range, i.e., from near-zero to infinite stiffness theoretically, ii) high-speed stiffness modulation using a relatively slow motor, and iii) zero/near-zero energy consumption for holding/altering the stiffness at equilibrium positions, and low-energy-cost stiffness modulation at non-equilibrium positions. Moreover, the proposed VSSEA has no limitation in motion range. To the best of our knowledge, the existing VSAs have yet to combine all these desired features of our proposed VSSEA [30].

The rest of the paper is organised as follows. In Section II, the mechanical design of the VSSEA is presented. In Section III, the dynamic model of the actuator is derived by using the analogy of a mass-spring-damper system and Euler-Bernoulli beam theory. In Section IV, the performance of the VSSEA is experimentally verified. The paper ends with discussion and conclusion given in Sections V and VI.

## II. MECHANICAL DESIGN

### A. Variable Stiffness Series Elastic Actuator:

Figure 1 illustrates the CAD model and the first prototype of the VSSEA. It comprises i) a conventional rigid actuator that includes a servo motor and a gearbox, illustrated by M1, ii) a novel variable stiffness actuation mechanism illustrated by VSAM, and ii) a direct drive servo motor illustrated by M2 in the figure.

The conventional rigid actuator M1 is used to independently control the equilibrium position of the output link. The proposed modular design approach enables us to freely tune the output torque and speed of the VSSEA. For example, we used Maxon EC90 flat motor and a 1:100 ratio harmonic drive to achieve ~100Nm output torque and ~π/2rad/s output speed in the first prototype of the actuator. The output power of the VSSEA can be directly adjusted by employing a different servo motor and/or a gearbox in the design of the conventional rigid actuator M1.

The second motor M2 is used to control the stiffness of the actuator via the VSAM independently. The low-energy-cost stiffness modulation feature, which is explained in Section III, of the VSAM allowed us to use the Maxon EC60 direct drive flat motor in the stiffness control of the actuator.

As shown in Fig.1, the proposed VSSEA is simply designed by integrating the VSAM to the conventional rigid actuator M1. This modular design approach provides great flexibility in building a VSA for different robotic applications. Let us now present the mechanical design of the novel VSAM that provides important features, such as a wide range of stiffness modulation and energy efficiency, in compliant actuation.

### B. Variable Stiffness Actuation Mechanism:

Figure 2 illustrates the CAD model and the first prototype of the VSAM. The design comprises i) eight radially distributed in-parallel leaf springs, ii) two rollers for each leaf spring to reduce friction and improve energy efficiency, and iii) a ball screw mechanism to move the rollers along the leaf springs as illustrated in the figure.

The stiffness of the actuator is modulated by changing the position of the rollers which is controlled by a ball-screw mechanism that is driven by the second motor M2. The VSSEA is in the softest mode when the rollers are at the free ends of the leaf springs, and the stiffness of the actuator increases as the rollers move towards the fixed end. With the nonlinear dynamic behaviour of the VSAM, a wide range of stiffness modulation can be obtained by simply changing the effective length of the leaf springs through the position control of the rollers. This allows us to perform large stiffness modulations within a short time, and this feature can provide several benefits in robotic applications [27]. Another important feature of the proposed VSSEA is low-energy-cost stiffness modulation. For example, the VSAM does not require constant power drain to hold the stiffness constant at equilibrium positions.

## III. DYNAMIC MODEL

The dynamic model of the VSSEA is obtained by employing the analogy of a mass-spring-damper system and the Euler-Bernoulli beam theory.

### A. Variable Stiffness Series Elastic Actuator:

The dynamic model of the VSSEA is illustrated in Fig. 3. In this figure, $J_\bullet$ represents the inertia of motor 1, gearbox, motor 2, and output link when • is $m1$, $g$, $m2$ and $l$, respectively; $\tau_\bullet$, $b_\bullet$, and $q_\bullet$ similarly represent the torque, viscous friction coefficient, and angle of the motor 1, gearbox, motor 2, and

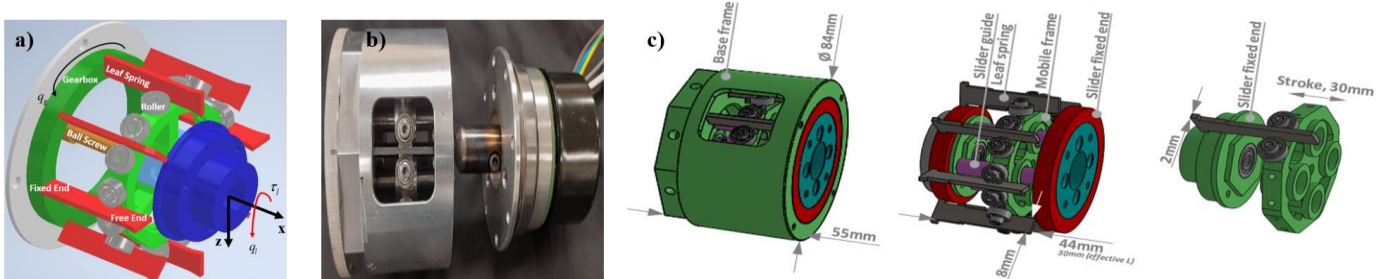

a) CAD model of leaf springs and mobile frame of the VSAM. b) First prototype of the VSAM. c) Dimensions of the VSAM.
Figure 2: CAD model and first prototype of the VSAM.

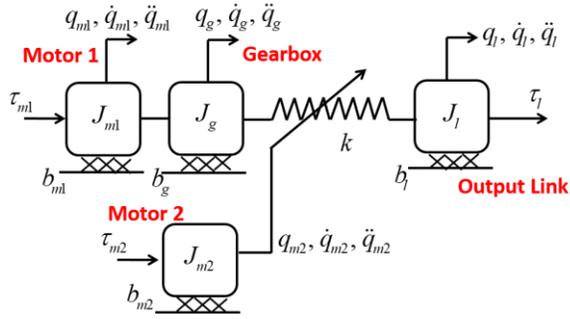

Figure 3: Dynamic model of the VSSEA.

output link, respectively; $\dot{q}_\bullet$ and $\ddot{q}_\bullet$ represent the first and second order derivatives of $q_\bullet$, i.e., angular velocity and angular acceleration, respectively; and $k$ represents the stiffness of the actuator, which can be defined as a nonlinear function of $q_{m2}$ and $q_l$ as shown in Section III.C.

While the first motor denoted by *m1* in Fig. 3 is used to control the equilibrium position of the actuator, the second motor denoted by *m2* is used to independently modulate the stiffness of the output link. The dynamic model of the VSSEA can be derived from this figure as follows:

$$\begin{aligned}\tilde{J}_{m1}\ddot{q}_{m1} + \tilde{b}_{m1}\dot{q}_{m1} &= \tau_{m1} - N^{-1}\tau_s - \tau_{m1}^{dis} \\ J_l\ddot{q}_l + b_l\dot{q}_l &= \tau_s - \tau_l - \tau_l^{dis} \\ J_{m2}\ddot{q}_2 + b_{m2}\dot{q}_{m2} &= \tau_{m2} - \tau_s^{dis} - \tau_{m2}^{dis}\end{aligned} \quad (1)$$

where $\tau_\bullet^{dis}$ represents the unknown/unmodelled disturbances of motor 1, motor 2, and output link when $\bullet$ is *m1*, *m2* and *l*, respectively; $\tau_s$ represents the torque exerted by the nonlinear springs of the VSAM on motor 1 and output link; $\tau_s^{dis}$ represents the disturbance torque exerted by the nonlinear springs of the VSAM on motor 2 in stiffness modulation; and $\tilde{J}_{m1} = J_{m1} + N^{-2}J_g$ and $\tilde{b}_{m1} = b_{m1} + N^{-2}b_g$ where *N* is gear ratio.

Equation (1) gives a simple yet useful dynamic model for the proposed VSSEA. However, more effort should be expended on understanding the nonlinear dynamics of the VSAM, i.e., deriving $\tau_s$ and $\tau_s^{dis}$ in Eq. (1). This will enable us to explain the important features such as the energy efficiency of the VSSEA.

### B. Variable Stiffness Actuation Mechanism:

To derive the model of the VSAM, let us focus on the first leaf spring illustrated in Fig. 4a. In this figure, $\mathbf{F}_{x_1}$, $\mathbf{F}_{y_1}$, and $\mathbf{F}_{z_1}$ represent the forces exerted by the leaf spring on the roller along the $\mathbf{x}_1$, $\mathbf{y}_1$, and $\mathbf{z}_1$ axes of the local coordinate frame on the leaf spring, respectively. It is noted that the proposed analysis can be similarly applied to the other leaf springs, e.g., the second leaf spring illustrated in Fig 4b.

When it is assumed that only the first leaf spring is used in the design of the VSAM, the kinematic and static equilibrium equations of the output link can be directly obtained from Fig. 4c and Fig. 4d as follows:

$$\delta = 2r\sin(q_l/2) \quad (2)$$

$$\tau_s - \tau_l = F_{r_1}r - \tau_l = \sqrt{F_{y_1}^2 + F_{z_1}^2}\, r - \tau_l = 0 \quad (3)$$

where $\delta$ is a kinematic constraint that relates the angle of the output link to the deflection of the leaf spring; *r* represents the

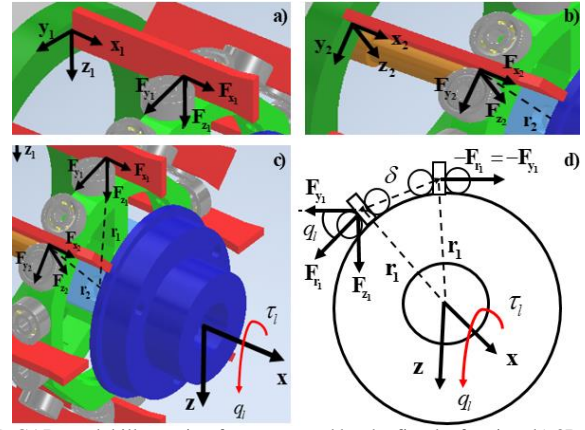

a) 3D CAD model illustrating forces exerted by the first leaf spring. b) 3D CAD model illustrating forces exerted by the second leaf spring. c) 3D CAD model illustrating the relation between the output torque and forces exerted on the first and second leaf springs. d) 2D kinematic model illustrating the relation between the deflection of the first leaf spring and output link angle.
Figure 4: Model of the Variable Stiffness Actuation Mechanism.

magnitude of the distance vector $\mathbf{r}_1$ illustrated in Fig. 4c; and $F_\bullet$ represents the magnitude of the force vector $\mathbf{F}_\bullet$ in which $\bullet$ can be $r_1$, $y_1$ and $z_1$ as shown in Fig. 4d.

When the other leaf springs illustrated in Fig. 2a are considered, the static equilibrium equation of the output link can be obtained by simply expanding Eq. (3) as follows:

$$\tau_s - \tau_l = \sum_{i=1}^{8} F_{r_i} r - \tau_l = \sum_{i=1}^{8} \sqrt{F_{y_i}^2 + F_{z_i}^2}\, r - \tau_l = 0 \quad (4)$$

where $F_{r_i}$, $F_{y_i}$, and $F_{z_i}$ similarly represent the forces exerted by the $i^{th}$ leaf spring, and *r* represents the magnitude of the distance vector $\mathbf{r}_i$. Equation (4) shows that the leaf spring forces should be identified to obtain the dynamic model of the VSSEA.

### C. Leaf Springs:

The leaf springs of the proposed VSAM not only bend along the lateral axes $\mathbf{y}_i$ but also slightly rotate about the longitudinal axes $\mathbf{x}_i$ at non-equilibrium positions [43]. For the sake of simplicity, the 3D deflection model of the leaf spring illustrated in Fig. 5 is numerically obtained using Finite Element Method

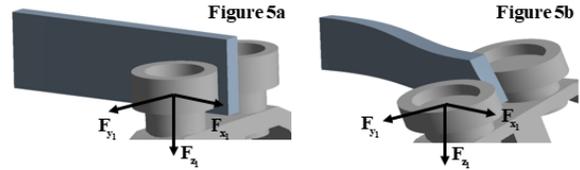

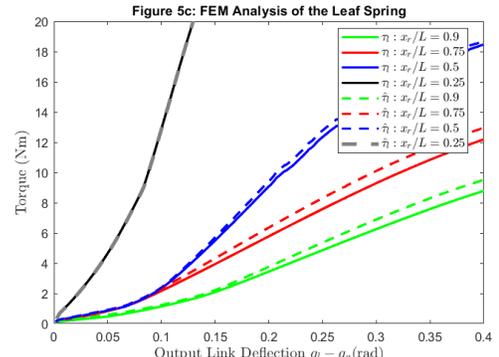

a) Beam at rest. b) Deflected beam c) Static output torque versus deflection. $\tau_l$ is the output torque and $\hat{\tau}_l = F_{y_1} r$ is the output torque associated with $\mathbf{F}_{y_1}$.
Figure 5: 3D large deflection model of the first leaf spring using FEM.

(FEM). While Fig. 5a shows the FEM model at an equilibrium position, Fig. 5b shows the maximum deflection of the leaf spring when the output link rotates for about 20°. As shown in this figure, the largest deflection occurs along the lateral axis $y_1$. Figure 5c illustrates the static output torque of the actuator for different stiffness configurations when the deflection of the output link increases. In this figure, $\hat{\tau}_l = F_{y_1} r$ represents the output torque associated with the force along the lateral axis at different non-equilibrium positions. It is clear from Fig. 5c that the output torque is mainly dominated by the bending force along the lateral axis within the deflection range of the actuator. This allows us to simplify the dynamic model of the VSAM while precisely estimating the output torque within a large deflection range.

To this end, let us consider the 2D deflection model of the leaf spring illustrated in Fig. 6. In this figure, $x_1$ and $y_1$ represent the position of a point on the beam along the $x_1$ and $y_1$ axes, respectively; $S_{x_1}$ and $\phi_{x_1}$ represent the arc length and slope of the beam at $x_1$, respectively; $\delta_{x_1}$ and $\delta_{y_1}$ represent the deflection of the beam along the $x_1$ and $y_1$ axes at $x_1$, respectively; $\delta_{x_r}$ and $\delta_{y_r}$ similarly represent the deflections at $x_r$ where the rollers are positioned on the $x_1$ axis; $L$ represents the length of the beam; and $\phi_{x_r}$ represents the slope of the beam at $x_r$.

The model of the leaf spring at a non-equilibrium position can be derived by using the nonlinear Euler-Bernoulli beam theory and Fig. 6 as follows.

$$S_{x_1} = \sqrt{\frac{EI}{2F_{y_1}}} \int_0^{\phi_{x_1}} \frac{d\phi}{\sqrt{\sin(\phi_{x_r}) - \sin(\phi)}} \quad (5)$$

$$x_1 = \sqrt{\frac{2EI}{F_{y_1}}} \int_0^{\phi_{x_1}} \frac{\cos(\phi) d\phi}{\sqrt{\sin(\phi_{x_r}) - \sin(\phi)}} \quad (6)$$

$$y_1 = \sqrt{\frac{2EI}{F_{y_1}}} \int_0^{\phi_{x_1}} \frac{\sin(\phi) d\phi}{\sqrt{\sin(\phi_{x_r}) - \sin(\phi)}} \quad (7)$$

where $E$ represents Young's modulus, and $I$ represents the moment of inertia of the beam cross section about the neutral axis [44]. The other parameters are same as defined earlier.

Equations (5 – 7) cannot be solved analytically so a numerical solution method should be employed to calculate the deflection of the beam for the applied force $F_{y_1}$. This allows us to calculate the reaction force exerted by the leaf spring on the roller, which is required for output torque calculation in Eq. (3). For example, the following steps can be employed to calculate the deflections of the leaf spring at $x_r$: i) Using a numerical method (e.g., Nonlinear shooting method or Adomian decomposition method) or the fzero command of Matlab, find $\phi_{x_r}$ in Eq. (5), and ii) Calculate $\delta_{x_r}$ and $\delta_{y_r}$ by numerically integrating Eqs. (6) and (7) over [0, $\phi_{x_r}$]. The kinematic and static equilibrium equations of the output link can be calculated by applying $\delta_{y_r} = \delta \cos(q_l)$ and $F_{y_1} \cong F_{r_1}$ to Eq. (2) and Eq. (3).

Although Eqs. (5-7) allows us to precisely estimate the output torque of the VSSEA for different deflection angles at non-equilibrium positions, they fall-short when attempting to explain how the output torque and stiffness of the actuator change as a function of $q_l$ and $q_{m2}$. Moreover, they are not very useful in controller analysis and synthesis. To tackle this problem, let us obtain an approximate model that provides a clear insight into the stiffness modulation characteristics of the proposed VSSEA.

When it is assumed that the deflections of the leaf springs are small, the torque and stiffness of the output link can be calculated using simple beam theory and Eqs. (2 – 4) as follows:

$$\tau_s(q_{m2}, q_l) = \frac{48EI}{\eta^3 q_{m2}^3} r^2 \sin\left(\frac{q_l}{2}\right) \quad (8)$$

$$k(q_{m2}, q_l) = \frac{24EI}{\eta^3 q_{m2}^3} r^2 \cos\left(\frac{q_l}{2}\right) \quad (9)$$

where $\eta = \ell/2\pi$ in which $\ell$ represents the lead of the ball screw, i.e., $x_r = \eta q_{m2}$ [30, 33, 44].

Equations (8) and (9) can be used to explain how the output torque and stiffness of the actuator change as a function of $q_l$ and $q_{m2}$. Equation (8) shows that the torque exerted by the nonlinear springs of the VSAM on the output link and motor 1 becomes higher as the angle of the output link increases. This occurs because larger deflections of the leaf springs lead to higher reaction forces exerted on the rollers. When the non-equilibrium position of the actuator remains constant, $\tau_s$ can still be regulated using the second motor. Equation (8) shows that as $q_{m2}$ decreases (increases), the torque exerted by the VSAM gets higher (lower). This occurs because the stiffness of the VSAM changes by the angle of the second motor (i.e., position of the roller) as shown in Eq. (9). The nonlinear behaviour of the VSAM allows us to obtain a wide range of stiffness modulation by simply controlling the angle of the second motor.

Figure 7 illustrates the torque and stiffness of the output link at different non-equilibrium positions. The accuracy of the small deflection model deteriorates as the angle of the output link increases. This is expected because this approximate model is obtained by assuming small deflections for the leaf springs. Despite some inaccuracies, the proposed approximate model is very useful to explain the important features of the VSSEA because the small and large deflection models exhibit similar behaviours as shown in Fig. 7. However, Eqs. (5 – 7) should be used to accurately estimate the output torque of the VSSEA.

At an equilibrium position where $q_l = q_g$, i) $\mathbf{F}_{y_1} = \mathbf{F}_{x_1} = \mathbf{0}$ when the stiffness of the actuator remains constant, i.e., $\dot{q}_{m2} = 0$, and ii) $\mathbf{F}_{y_1} = \mathbf{0}$, but $\mathbf{F}_{x_1} \cong \mathbf{0}$ due to small friction and inertial forces when $\dot{q}_{m2} \neq 0$. Therefore, the energy consumption of stiffness modulation is negligible at equilibrium positions.

The energy consumption of stiffness modulation increases as the deflections of the leaf springs get larger at non-equilibrium

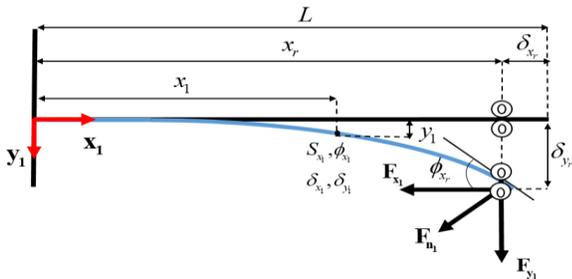

Figure 6: 2D large deflection model of the first leaf spring.

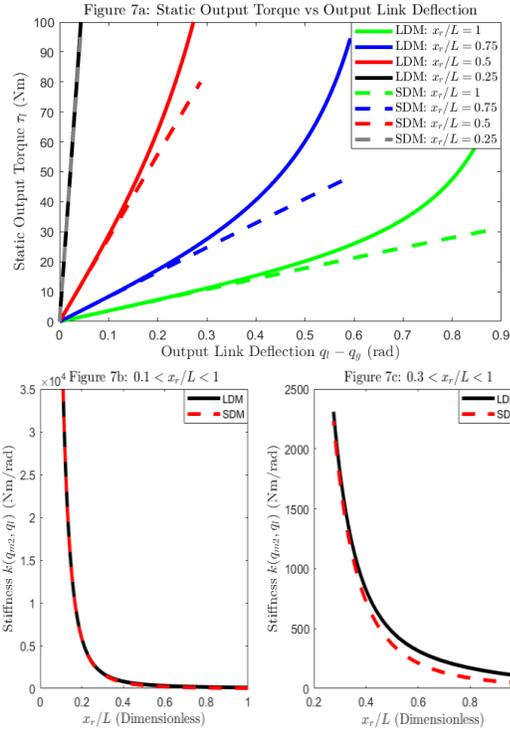

a) Static output torque versus deflection. b) 90% stiffness modulation. c) 70% stiffness modulation.

Figure 7: Output torque and stiffness of the VSSEA. LDM: Large Deflection Model and SDM: Small Deflection Model.

positions. The disturbance torque exerted by the VSAM on the stiffness modulation motor can be calculated as follows:

Large Deflection Model: $\tau_s^{dis} = \sum_{i=1}^{8} F_{x_i} r$ (10)

Small Deflection Model: $\tau_s^{dis} = \frac{48EI}{\eta^3 q_{m2}^3} r^2 \sin\left(\frac{q_l}{2}\right)\tan(\varphi)$ (11)

where $\varphi = F_y L^2 / 2EI$ [44].

Figure 8 illustrates the disturbance torque of the stiffness modulation motor $\tau_s^{dis}$ and the static output torque of the VSSEA $\tau_s = \tau_l$ when the deflection of the output link increases. Since $\tau_s^{dis}$ is zero when $q_l = q_g$, the VSAM does not consume energy to keep the stiffness constant at equilibrium positions. When the deflection is small, the energy consumption of the VSAM is negligible because $\tau_s^{dis}$ is very low regardless of $\tau_s$. However, the VSAM consumes higher energy as the deflection of the output link increases. It is clear from Fig. 8 that the disturbance torque of the stiffness modulation motor is always limited within the work-range of the VSSEA. This disturbance can be further supressed using novel mechanical designs, e.g., in [45].

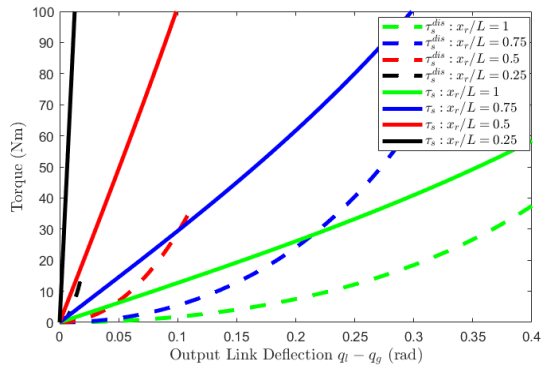

Figure 8: Disturbance torque of the stiffness modulation motor.

## IV. MOTION CONTROL OF THE VSSEA

In this section, the position and force control problems of the VSSEA are discussed. It is a well-known fact that the motion control problem of compliant actuators is more complicated than that of conventional stiff actuators, particularly in position control [10]. To precisely track link trajectories or interact with unstructured environments, internal and external disturbances should be compensated using adaptive and robust controllers [9 – 15]. This section experimentally shows that the proposed VSSEA allows us to conduct high-performance motion control tasks using conventional PID controllers. The experimental setup was built by using ESCON 50/5 motor drivers, 1000ppr encoders at the motors and a 10000ppr encoder at the link. A PC with a Linux operating system was employed to perform the real-time motion control experiments with 1ms sampling time.

### A. Position Control:

Let us start with the position control problem of the VSSEA. Figure 9 illustrates the position control experiments when a PID controller is synthesised for controlling the angle of the first motor as follows:

$$\tau_m = K_p\left(q_g^{ref} - q_g\right) + K_i\int\left(q_g^{ref} - q_g\right)dt + K_d\left(\dot{q}_g^{ref} - \dot{q}_g\right)$$ (12)

where $q_g^{ref}$ and $\dot{q}_g^{ref}$ represent the position and velocity references of the gearbox, i.e., $q_g$, respectively.

External disturbances up to 5Nm and 15Nm were applied to the output link after 2 seconds when the VSSEA was in soft (21 Nm/rad) and stiff mode (985 Nm/rad), respectively. Although the transient response was good except a relatively high overshoot at link side, the link of the actuator was very sensitive to external disturbances, particularly when the actuator was in soft mode. This is expected because the angle of the link $q_l$ is not used in the position controller synthesis.

To improve the robustness against external load, let us use $q_l$ in the feedback controller synthesis. In this experiment, the control signal was designed as follows:

$$\tau_m = K_p\left(q_l^{ref} - q_l\right) + K_i\int\left(q_l^{ref} - q_l\right)dt + K_d\left(\dot{q}_l^{ref} - \dot{q}_l\right)$$ (13)

where $q_l^{ref}$ and $\dot{q}_l^{ref}$ similarly represent the position and velocity references of the VSSEA's link, respectively.

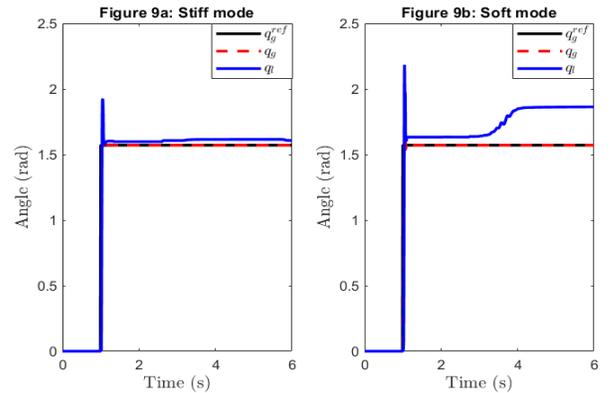

a) Stiff mode. b) Soft mode.
Figure 9: Position regulation control of motor 1 when external loads are applied to the link. $K_p = 15000$, $K_d = 500$, $K_i = 75$ and $q_g^{ref} = 0.5\pi$.

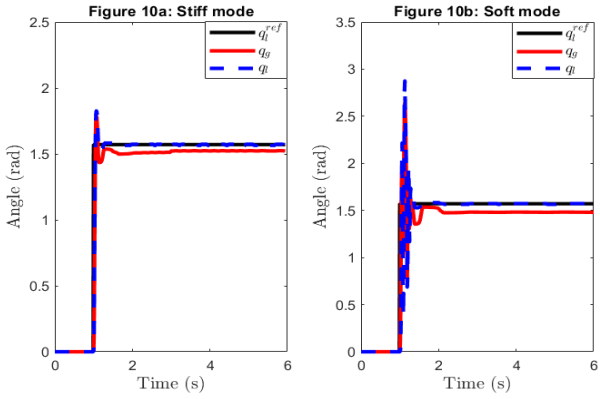

a) Stiff mode. b) Soft mode
Figure 10: Position regulation control of the output link when external loads are applied to the link. $K_p = 5000$, $K_d = 95$, $K_i = 35$, and $q_l^{ref} = 0.5\pi$.

External disturbances up to 10Nm were similarly applied when the actuator was in stiff mode, however only less than 1Nm external disturbances could be applied due to robust stability problems encountered in the soft mode of the actuator. Figure 10 shows that the robustness against external loads was improved by using $q_l$ in the feedback controller synthesis. However, the performance of transient response was notably degraded by large vibrations when the actuator was in soft mode (see Fig. 10b). Moreover, only small external disturbances could be suppressed due to the robust stability problems. It is a well-known fact that more advanced controllers should be employed for the robust position control problem of compliant actuators [10]. By changing the stiffness of the actuator, this paper proposes a simple yet effective solution for this challenging problem as shown in Fig. 10.

When we used the same PID controller in trajectory tracking control, we obtained similar results as illustrated in Fig. 11. Due to the well-known bandwidth limitations of compliant systems, increasing the speed of reference trajectory notably degraded the position control performance when the actuator was in soft mode [3]. The VSSEA could not follow 1Hz reference trajectory as illustrated in Fig. 11b. The performance of trajectory tracking control could be easily improved by increasing the stiffness of the actuator as illustrated in Figs. 11c and 11d.

*B. Force Control:*

By using Hooke's law, the force control problem of the compliant actuator was described as a position control problem, and force control experiments were performed by controlling the deflection of the output link at non-equilibrium positions. Similar to the position control experiments, a simple PID controller was synthesised by feeding back the deflection angle of the output link, i.e., $\Delta_q = q_l - q_g$ where $q_l$ and $q_g = q_{m1}/N$ are the link and gear angles, respectively.

Figure 12 illustrates the force control experiments of the VSSEA. The force control signal was designed using Eq. (14).

$$\tau_m = K_p\left(\Delta_q^{ref} - \Delta_q\right) + K_i\int\left(\Delta_q^{ref} - \Delta_q\right)dt + K_d\left(\dot{\Delta}_q^{ref} - \dot{\Delta}_q\right) \quad (14)$$

where $\Delta_q^{ref}$ and $\dot{\Delta}_q^{ref}$ represent the position and velocity references of the link deflection, i.e., $\Delta_q = q_l - q_g$, respectively.

When the actuator was in soft mode, relatively large link deflections occurred for small output torques as shown in Figs. 12a and 12b. This allows actuator to physically interact with different environments in a safe manner. To achieve higher

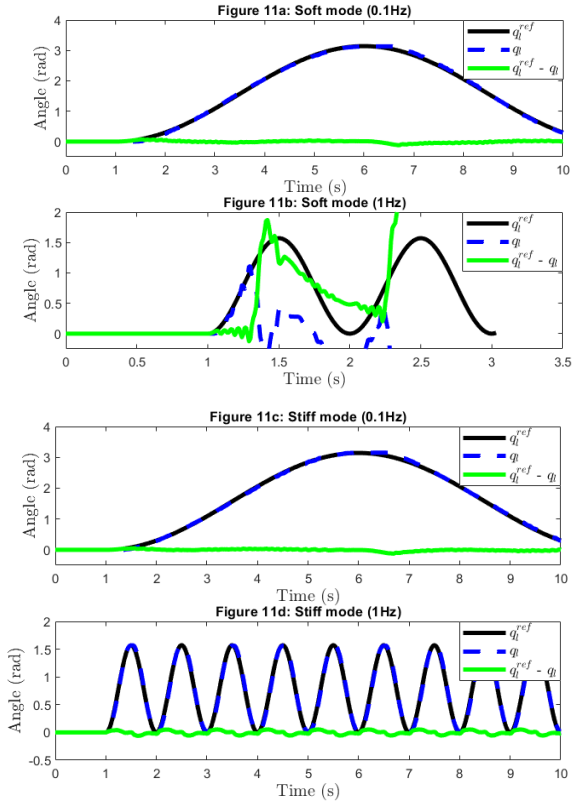

a) Soft mode and $f = 0.1$Hz b) Soft mode and $f = 1$Hz. c) Stiff mode and $f = 0.1$Hz. d) Stiff mode and $f = 1$Hz.
Figure 11: Position trajectory tracking control of the output link when $K_p = 5000$, $K_d = 95$, $K_i = 35$, and $q_l^{ref} = K\left(1 - \cos\left(2\pi f(t-1)\right)\right)$ where $K$ is $0.5\pi$ and $2\pi$.

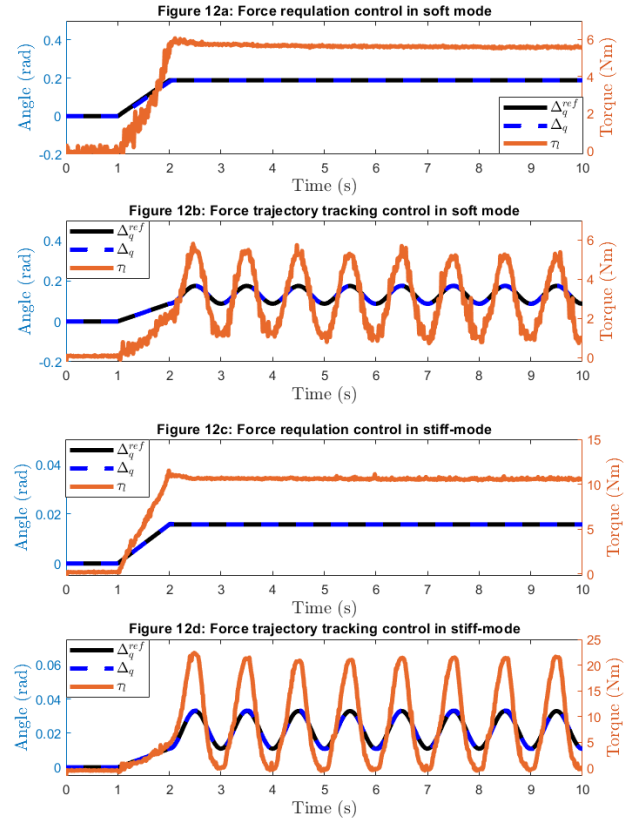

a) Regulation control in soft mode b) Trajectory tracking control in soft mode c) Regulation control in stiff mode d) Trajectory tracking control in stiff mode.
Figure 12: Force control experiments when $K_p = 2500$, $K_d = 85$, and $K_i = 15$.

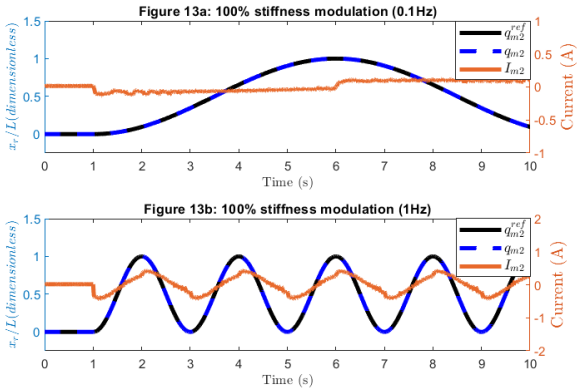

a) 0.1Hz stiffness modulation. b) 1Hz stiffness modulation.
Figure 13: Stiffness modulation at an equilibrium position.

output torque, the stiffness of the actuator should be increased as illustrated in Figs. 12c and 12d. This, however, may degrade safety in contact motion. It is clear from Fig. 12 that the proposed VSSEA enables us to conduct high-performance force control applications using a conventional PID controller.

*C. Stiffness Modulation:*

Figure 13 illustrates the position control experiments of motor 2 at different speeds when the actuator is at equilibrium positions. The motor did not draw current, i.e., the VSAM did not consume energy, to keep the stiffness of the actuator fixed at an equilibrium position as shown in Fig. 13. The current of motor 2 was negligible at low speeds of stiffness modulation, e.g., 0.1Hz in Fig. 13a. As the speed of stiffness modulation increased, the motor drew higher current due to the increased frictional and inertial disturbances of the VSAM (see Fig. 13b). This is expected because the energy consumptions of all variable stiffness actuators become higher as the speed of stiffness modulation is increased [33, 45]. Since the frictional and inertial disturbances were mitigated in the design of the VSAM, the stiffness modulation could be performed using low current drains at equilibrium positions as shown in Fig. 13.

Figure 14 illustrates the stiffness modulation experiments when the actuator is at non-equilibrium positions. While the deflection angle of the output link was regulated, the stiffness of the actuator was modulated 10% in this experiment. Figures 14a and 14b show that the current of motor 2, $I_{m2}$, was low when 10% stiffness modulation was performed in the stiff mode of the VSSEA. This is expected because, as shown in Eq. (11), the small deflections of leaf springs in stiff mode could lead to low disturbance torques exerted by the VSAM on motor 2. Compared to other energy efficient and infinite-range variable stiffness actuators, e.g., [34, 35], this feature of the proposed VSAM allows us to perform infinite-range stiffness modulation using bounded control signals [33]. Figures 14c and 14d show that the current drain of motor 2 was higher when 10% stiffness modulation was performed in the soft mode of the VSSEA. Since the deflection angle of the output link was larger in soft mode, the current drain of motor 2 increased by the higher disturbance torque of the VSAM as shown in Eq. (11). Figures 13 and 14 show that the control signal of motor 2 was always bounded when stiffness modulations were conducted at equilibrium and non-equilibrium positions.

Figure 15 illustrates the energy cost of changing stiffness compared to the potential energy stored in the springs of the VSAM when 10% stiffness modulation was performed at stiff and soft modes of the actuator. It shows that the energy cost of

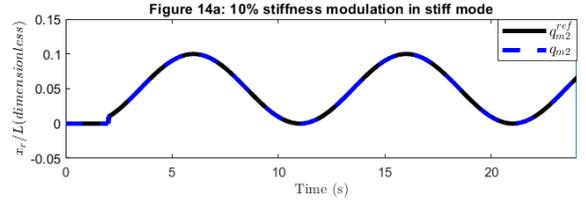
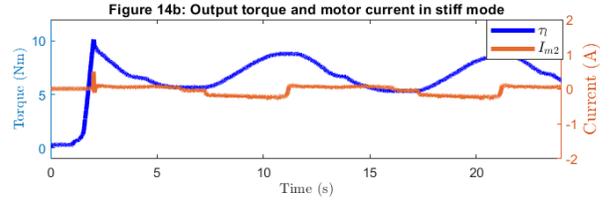
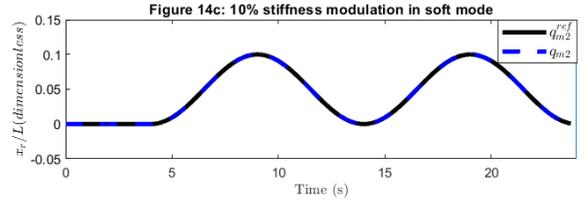
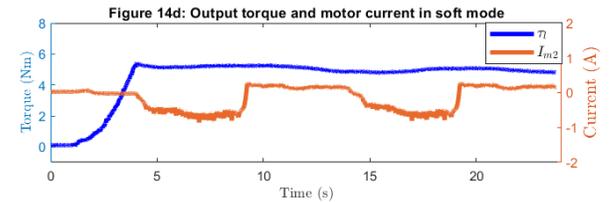

a) Stiffness modulation in stiff mode. b) Output torque and motor current in stiff mode. c) Stiffness modulation in soft mode. d) Output torque and motor current in soft mode.
Figure 14: Stiffness modulation at non-equilibrium positions.

stiffness modulation is weakly coupled to the deflection of the output link. While lower energy was consumed by the VSAM at fixed stiffness configurations in both stiff and soft modes, the energy cost of stiffness modulation increased when the stiffness of the actuator was changed. The larger deflection of the output link in soft mode led to higher energy cost of stiffness modulation as shown in Fig. 15b. This result is expected because the disturbance torque gets larger as the deflection of the output link increases as shown in Eq. (11).

## V. DISCUSSION

The experimental results given in Section IV show that the

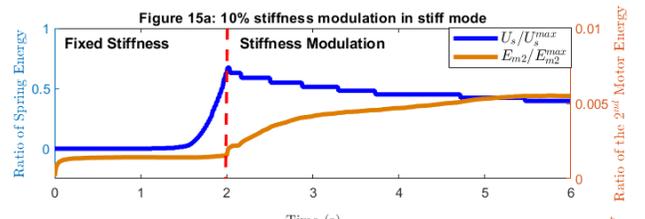
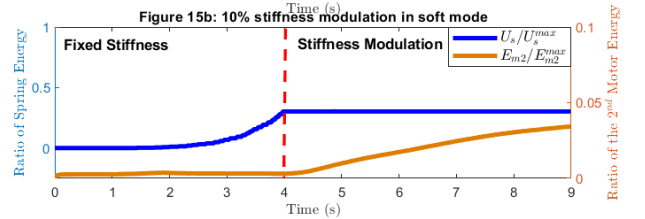

a) Stiffness modulation in stiff mode. b) Stiffness modulation in soft mode.
Figure 15: Energy consumption of the VSAM.

proposed VSSEA enables us to conduct high-performance position and force control applications using conventional PID controllers. This provides several benefits in many different advanced robotic applications. For example, while the stiff mode of the VSSEA allows a cobot to perform high-precision position control tasks in industry, the soft mode of the actuator can boost safety in physical robot-environment interaction. However, more advanced controllers should be synthesised for smooth transition between position and force control tasks, i.e., the stiff and soft mode of the actuator. In addition, more effort should be expended on optimal stiffness modulation.

The experimental results show that a wide range of stiffness modulation can be achieved by simply controlling the position of the second motor. For example, while the stiffness of the output link is 21Nm/rad when the actuator is in soft mode in Fig. 9a, it is ~50 times higher with 985Nm/rad in Fig. 9b. This is an important feature of the proposed VSSEA as it allows us to perform both precise position control and safe robot-environment interaction tasks using conventional PID controllers. Other important features of the VSSEA are fast and low-energy-cost stiffness modulation capabilities. The transition from the softest mode to the stiffest mode can be achieved within a second as illustrated in Fig. 13. Since the disturbance torque exerted by the VSAM on motor 2 is always bounded unless it is zero, a wide range of stiffness modulation can be performed using bounded control signals at equilibrium and non-equilibrium positions as shown in Figs. 13 and 14. Therefore, the energy cost of stiffness modulation is low as shown in Fig. 15. This can provide significant benefits to mobile robotic systems such as humanoids, quadrupeds, and exoskeletons.

With the proposed modular design approach, the VSSEA can be easily modified to meet the requirements of different robotic applications. For instance, i) torque density can be increased using a higher gear ratio, ii) stiffness characteristic can be tuned using the different number, material, and shape of leaf springs, and iii) faster stiffness modulation can be achieved using a ball screw with higher lead. The proposed novel mechanical design eliminates the work-range limitation of variable stiffness actuators as shown in Figs. 11a and 11c. This allows us to apply the VSSEA to various robotic applications.

By neglecting the small torsional motions of leaf springs, the dynamic model of the VSSEA is obtained using the analogy of a mass-spring-damper system and the Euler-Bernoulli beam theory in Section III [44]. While the exact dynamic model and nonlinear Euler-Bernoulli beam theory are computationally expensive and ineffective in controller synthesis [43, 44], the model derived through simple beam theory is inaccurate when the deflection of the output link is large. Thus, more effort should be expended to obtain an effective dynamic model for the VSSEA.

## VI. CONCLUSION

This paper proposes a new VSSEA that can perform fast and low-energy-cost stiffness modulation over a large range. These features can provide several benefits to robotic systems such as easier motion control problems, longer battery life, and safer physical robot environment interaction. It is experimentally shown in this paper that the proposed VSSEA allows us to conduct high-performance position and force control tasks using conventional PID controllers. It is also demonstrated that the stiffness of the actuator can be increased or decreased up to 50 times within a second while the energy cost of stiffness modulation is very low. However, further research should be conducted to clarify how the VSSEA can contribute to robotic applications. To this end, we will apply the proposed actuator to advanced robotic applications such as legged locomotion in the future studies.

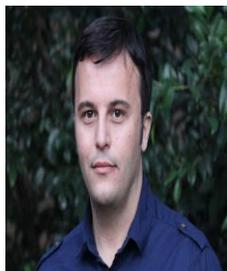

**Emre Sariyildiz** (S'11, M'16, SM'21) received his first Ph.D. degree in Integrated Design Engineering from Keio University, Yokohama, Japan, in 2014, and second PhD degree in Control and Automation Engineering from Istanbul Technical University, Istanbul, Turkey, in 2016.

He is currently a Senior Lecturer at the School of Mechanical, Materials, Mechatronic, and Biomedical Engineering, University of Wollongong, Wollongong, NSW, Australia. His research interests include control theory, robotics, mechatronics, and motion control.

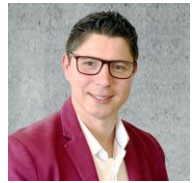

**Rahim Mutlu** (M'22) received his Ph.D. degree in Robotics from University of Wollongong, Wollongong (UOW), NSW, Australia, in 2013. He was a Lecturer with UOW 2017-21, prior to committing his current role as Assistant Professor with Faculty of Engineering and Information Sciences at UOWD, UAE. He is also founder of the Intelligent Robotics & Autonomous Systems Co (iR@SC), NSW, 2529, Australia. His research interests include soft robotics, soft haptics, wearable technologies, assistive and rehabilitation exoskeletons, additive manufacturing.

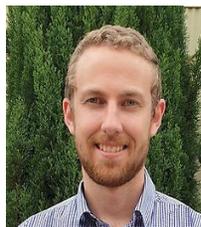

**Jon Roberts** received his Ph.D. degree in Mechanical Engineering from University of Wollongong (UOW), Wollongong, NSW, Australia in 2019.

Since January 2020, he has been a Lecturer with the School of Mechanical, Materials, Mechatronic, and Biomedical Engineering, University of Wollongong, Wollongong, NSW, Australia. His research interests are simulation methods, bulk materials handling, safety in mining, and dust control technology.

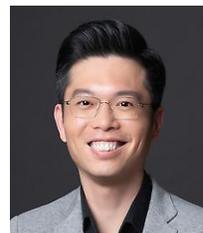

**Chin-Hsing Kuo** received his Ph.D. degree in Mechanical Engineering from King's College London, UK, in 2011.

Since February 2019, he has been a Senior Lecturer with the School of Mechanical, Materials, Mechatronic, and Biomedical Engineering, University of Wollongong, Wollongong, NSW, Australia. Before that he was an Associate Professor with the National Taiwan University of Science and Technology. His research interests include parallel robots, mechanism design, and robot kinematics and dynamics.

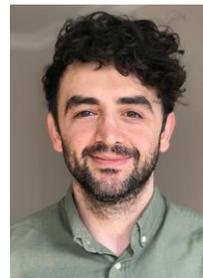

**Barkan Ugurlu** (S'08-M'10) received his Ph.D. degree in Electrical and Computer Engineering from Yokohama National University, Yokohama, Japan, in 2010.

He was a Marie Sklodowska-Curie Fellow and currently holds an Assistant Professor position at the Department of Mechanical Engineering, Ozyegin University, Istanbul, Turkey. His research interests include legged locomotion control, hardware development for novel robotic systems, and multi-body dynamics.